\documentclass[conference]{IEEEtran}
\IEEEoverridecommandlockouts
\usepackage{cite}
\usepackage{amsmath,amssymb,amsfonts}
\usepackage{algorithmic}
\usepackage{array}
\usepackage{graphicx}
\usepackage{longtable}
\usepackage{booktabs}
\usepackage{multirow}
\usepackage{textcomp}
\usepackage{soul}
\usepackage{xcolor}
\usepackage{svg}
\def\BibTeX{{\rm B\kern-.05em{\sc i\kern-.025em b}\kern-.08em
    T\kern-.1667em\lower.7ex\hbox{E}\kern-.125emX}}

\usepackage{fancyhdr}
\usepackage{geometry}
\usepackage{lipsum} % For dummy text

\begin{document}

\title{Accident-Driven Congestion Prediction and Simulation: An Explainable Framework Using Advanced Clustering and Bayesian Networks
}

%Accident-Driven Congestion Prediction and Simulation: An Explainable and Prescriptive Framework Using Advanced Clustering and %Bayesian Networks

\author{
\IEEEauthorblockN{
Kranthi Kumar Talluri\IEEEauthorrefmark{1}, Galia Weidl\IEEEauthorrefmark{1}, Vaishnavi Kasuluru\IEEEauthorrefmark{2} 
}

\IEEEauthorblockA{\IEEEauthorrefmark{1}Aschaffenburg University of Applied Sciences, Aschaffenburg, 63743 Germany.}
\IEEEauthorblockA{\IEEEauthorrefmark{2}Centre Tecnològic de Telecomunicacions de Catalunya (CTTC), Barcelona, 08860 Spain.}
\IEEEauthorblockA{Email: %\IEEEauthorrefmark{1}, 
\IEEEauthorrefmark{1}\{KranthiKumar.Talluri, Galia.Weidl\}@th-ab.de,
\IEEEauthorrefmark{2}kasuluru.vaishnavi@upc.edu}
}

\maketitle

\begin{abstract}

Traffic congestion due to uncertainties, such as accidents, is a significant issue in urban areas, as the ripple effect of accidents causes longer delays, increased emissions, and safety concerns. To address this issue, we propose a robust framework for predicting the impact of accidents on congestion. We implement Automated Machine Learning (AutoML)-enhanced Deep Embedding Clustering (DEC) to assign congestion labels to accident data and predict congestion probability using a Bayesian Network (BN). The Simulation of Urban Mobility (SUMO) simulation is utilized to evaluate the correctness of BN predictions using evidence-based scenarios. Results demonstrate that the AutoML-enhanced DEC has outperformed traditional clustering approaches. The performance of the proposed BN model achieved an overall accuracy of 95.6\%, indicating its ability to understand the complex relationship of accidents causing congestion. Validation in SUMO with evidence-based scenarios demonstrated that the BN model's prediction of congestion states closely matches those of SUMO, indicating the high reliability of the proposed BN model in ensuring smooth urban mobility.

\end{abstract}


\section{Introduction}
Urban transportation systems worldwide face a significant challenge due to traffic congestion caused by various factors, including inadequate infrastructure, severe environmental conditions, poorly coordinated traffic signals, and road accidents \cite{santos2021machine}. Congestion can be broadly categorized into two categories. When traffic patterns are regular and traffic demands are predictable, such as during morning rush hours and evening peak hours, this type of congestion is referred to as recurring congestion. On the other hand, if the nature of congestion is unpredictable, like climatic disasters, accidents, and extreme weather conditions, it can be categorized as non-recurring congestion \cite{afrin2021probabilistic}. Tackling and mitigating non-recurring congestion becomes challenging due to its non-volatile occurring patterns.
Accidents, in particular, are crucial to mitigate as they disrupt traffic flow and endanger human safety, most likely resulting in secondary incidents triggered by abrupt braking and bottleneck situations \cite{luan2022traffic}. Furthermore, they have a broader impact on quality of life as people tend to experience travel delays, loss of productivity, fuel wastage due to long waits in traffic queues, and annoyance caused by bottlenecks, promoting a need for better traffic management and road safety \cite{talluri2024bayesian}. The U.S. recorded 12 million crashes causing infrastructure damage, loss of life, and severe traffic congestion in the year of 2018 \cite{mudgil2022intensity}. The traffic accidents reported by China’s Public Security Ministry from 2009 to 2011 resulted in 65k fatalities and 250k injuries \cite{ManzoorRFCNN2021}. These instances highlight the impact of accidents and congestion on the economy and public safety. 

% \begin{figure*}[h]
%     \centering
%     \includesvg[width=5\textwidth, height=0.12\textheight]
%     {figures/WorkFlow.svg} 
%     \caption{Workflow of traffic congestion analysis}
%     \label{fig:WorkFlow} 
% \end{figure*}

% \begin{figure*}[h]
%     \centering
%     \resizebox{0.80\textwidth}{0.13\textheight}{
%         \includegraphics{figures/WorkFlow.png}
%     }
%     \caption{Workflow of traffic congestion framework.}
%     \label{fig:WorkFlow} 
% \end{figure*}

\begin{figure*}[h]
    \centering
    \includegraphics[width=\linewidth]{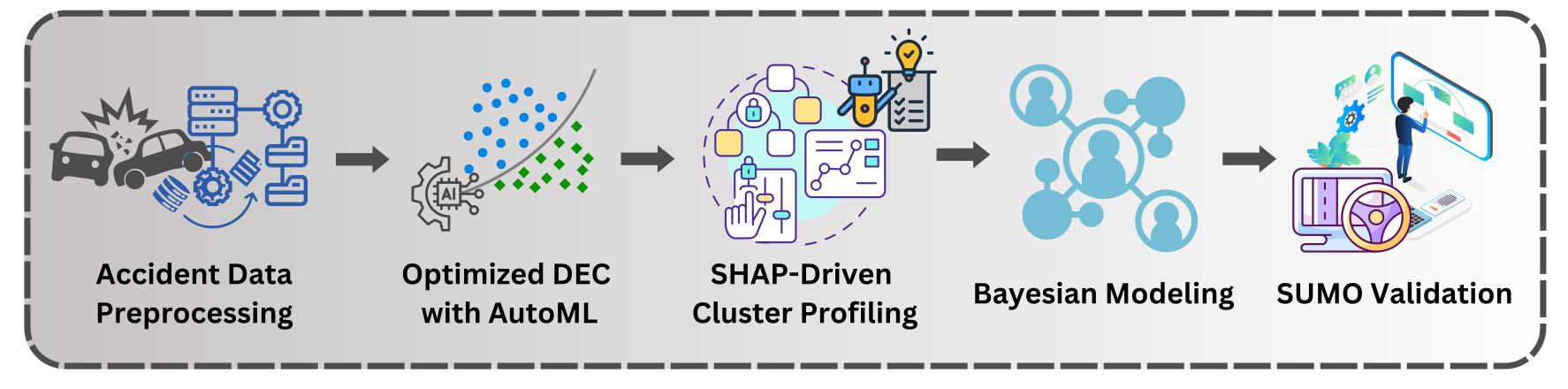} 
    \caption{Workflow of traffic congestion framework.}
    \label{fig:WorkFlow}
\end{figure*}

Many researchers have investigated accident data using traditional clustering techniques, such as k-means, hierarchy, and DBSCAN \cite{islam2021clustering}. However, conventional approaches fail to handle the accident data due to its high-dimensional, outlier-sensitive, and non-linear characteristics, resulting in suboptimal insights. These limitations affect the understanding of the hidden relationship between accidents and congestion. To overcome these challenges, we introduce an innovative framework that combines the insights of advanced clustering techniques \cite{zhou2023identifying, desai2023optimal} with probabilistic forecasting models to help extract the crucial insights necessary for planning effective congestion mitigation strategies, as shown in Fig. \ref{fig:WorkFlow}. A probabilistic Bayesian network provides a comprehensive understanding of the features that influence traffic congestion. Furthermore, to evaluate the performance of the BN, we have integrated it with the Simulation of Urban Mobility (SUMO) to bridge the gap between theoretical insights and practical applications. SUMO provides a realistic environment to validate the outcomes of BN for congestion predictions by simulating various accident scenarios, ensuring the framework's reliability and validating it through use cases in real-world deployments. The main key contributions of this work are as follows:

\begin{itemize}
    \item Deep Clustering with AutoML Optimization: Extracted high-quality clusters from accident data utilizing Deep Embedded Clustering (DEC) enhanced by the AutoML technique named Optuna. 
    \item Comparison with Traditional Methods: Demonstrated the superior performance of DEC over k-means, DBSCAN, and hierarchical clustering.
    \item Explainability with Shapley Values: Quantified feature importance to ensure transparency for BN modeling.
    \item BN for Congestion Prediction: Developed a probabilistic model to predict congestion probabilities and analyzed root causes through various scenarios.
    \item Simulation-Based Validation: Simulated evidence-based accidents in SUMO to validate BN predictions. 
\end{itemize}

The rest of the paper is organized as follows. Sections II and III provide an overview of related work and background information on clustering and BN. Followed by BN modeling, SUMO setup, and results in Sections IV and V.

\section{Related Works}

Traffic congestion and road safety play a vital role in transportation research. A comprehensive analysis of congestion caused by accidents in \cite{zheng2020determinants} revealed that 8.4\% more congestion was attributed to rear-end collisions compared to other types of accidents. The heavy vehicle accidents caused 6.03\% more congestion. Spatial effects on reduced traffic flow showed 28\% dependency on adjacent link congestion in Beijing's urban area \cite{zheng2024exploring}. Congestion, if not controlled at an early stage, has the potential to spread through urban networks. A novel threshold-based approach was developed in \cite{talluri2024impact}, utilizing these insights to correlate traffic incidents with congestion.

As accidents play a pivotal role in congestion, it is highly significant to analyze them using techniques like clustering, which are widely used to identify and analyze accident hotspots effectively. Traditional clustering approaches, such as k-means, DBSCAN, and hierarchical techniques, can help group accident data based on their patterns. However, research shows that DBSCAN is particularly effective at handling spatial traffic data and identifying hotspots, which are more prone to accidents \cite{islam2021clustering}. Integrating GIS-based spatial studies into clustering is advantageous in obtaining accident geographical trends along with their dependency on surrounding infrastructure \cite{kamh2024exploring}. Introducing ML-based analysis approaches can also be beneficial \cite{santos2021machine} for practical congestion analysis. Nevertheless, traditional clustering techniques struggle to handle high-dimensional data and are highly sensitive to the selection of appropriate parameters. Furthermore, they do not effectively capture complex non-linear relations of traffic patterns. Feature engineering and parameter tuning in traditional approaches need extensive domain expertise. To avoid such manual parameter selection, an advanced clustering approach integrated with deep learning aids in building more sophisticated clustering techniques, such as DEC. Authors in \cite{zhou2023identifying} demonstrated that DEC with stacked encoders performed efficient traffic pattern analysis and identified pre-crash scenarios. They proved to be more advantageous in feature extraction than traditional approaches.  Moreover, DEC achieved 95\% accuracy in finding accident hotspots and effectively handled the complex spatio-temporal patterns in the use cases of optimal ambulance positioning \cite{desai2023optimal}. DEC’s capabilities were further explored in \cite{han2019short} by predicting short-term traffic through the incorporation of convolutional neural networks with triplet loss. However, these research works didn’t explore the advantage of using AutoML methods, which further enhance the performance of deep learning-based methods through automated optimal hyperparameter tuning. Furthermore, various existing ML and deep learning techniques can be used to predict or perform accident analysis. However, these approaches lack explainability, which can be addressed by using Bayesian networks. 

BNs can achieve accurate predictions with modeled uncertainties and explainability. The probabilistic estimation framework built using the BN approach in \cite{afrin2021probabilistic} effectively distinguished recurring and non-recurring congestion scenarios. This field has been advanced in congestion propagation analysis using a dynamic Bayesian graph convolutional network, which has shown better accuracy in congestion prediction \cite{luan2022traffic}. BN provides effective uncertainty quantification and causal inference, and the validation of its performance through real-world simulation is crucial to ensure that the framework is robust, flexible, and scalable when deployed in real-world infrastructure. Furthermore, the AutoML and Bayesian reasoning principles adopted here align well with emerging closed-loop optimization strategies envisioned in 6G networks, as discussed in \cite{bazzi2025upper}.

%This work addresses all these limitations through its key contributions by enhancing the performance of DEC using AutoML. Moreover, a BN model is built to analyze and predict congestion, which is further validated using the SUMO simulator. 

This research aims to develop a robust framework and bridge the existing literature's gaps by combining advanced clustering, explainable AI, and simulation-based performance validation for effective decision-making.  

\section{Background Information}

\subsection{Dataset Description}
The accident dataset used in this work, obtained from the open-source repository, consists of 1.3 billion accidents across 49 states of the USA from February 2016 to March 2023 \cite{moosavi2019accident}. A subset of 50k records from the year 2022 is randomly selected to ensure a balanced and resource-efficient computation. The dataset encompasses a wide range of features related to accident characteristics, facilitating a detailed analysis of congestion caused by accidents. It is preprocessed, scaled, and encoded to ensure that no major accident types are excluded by stratifying samples across severity levels, timestamps, and locations before cleaning. This preserved representative diversity in the selected subset. This processed dataset is clean, consistent, and suitable for performing clustering. Moreover, the continuous features were discretized to help the BN model probabilistic relationships more effectively by simplifying the representation of dependencies and enabling clearer inference from categorical features.

\begin{figure}[h]
    \centering
    \includegraphics[width=\linewidth]{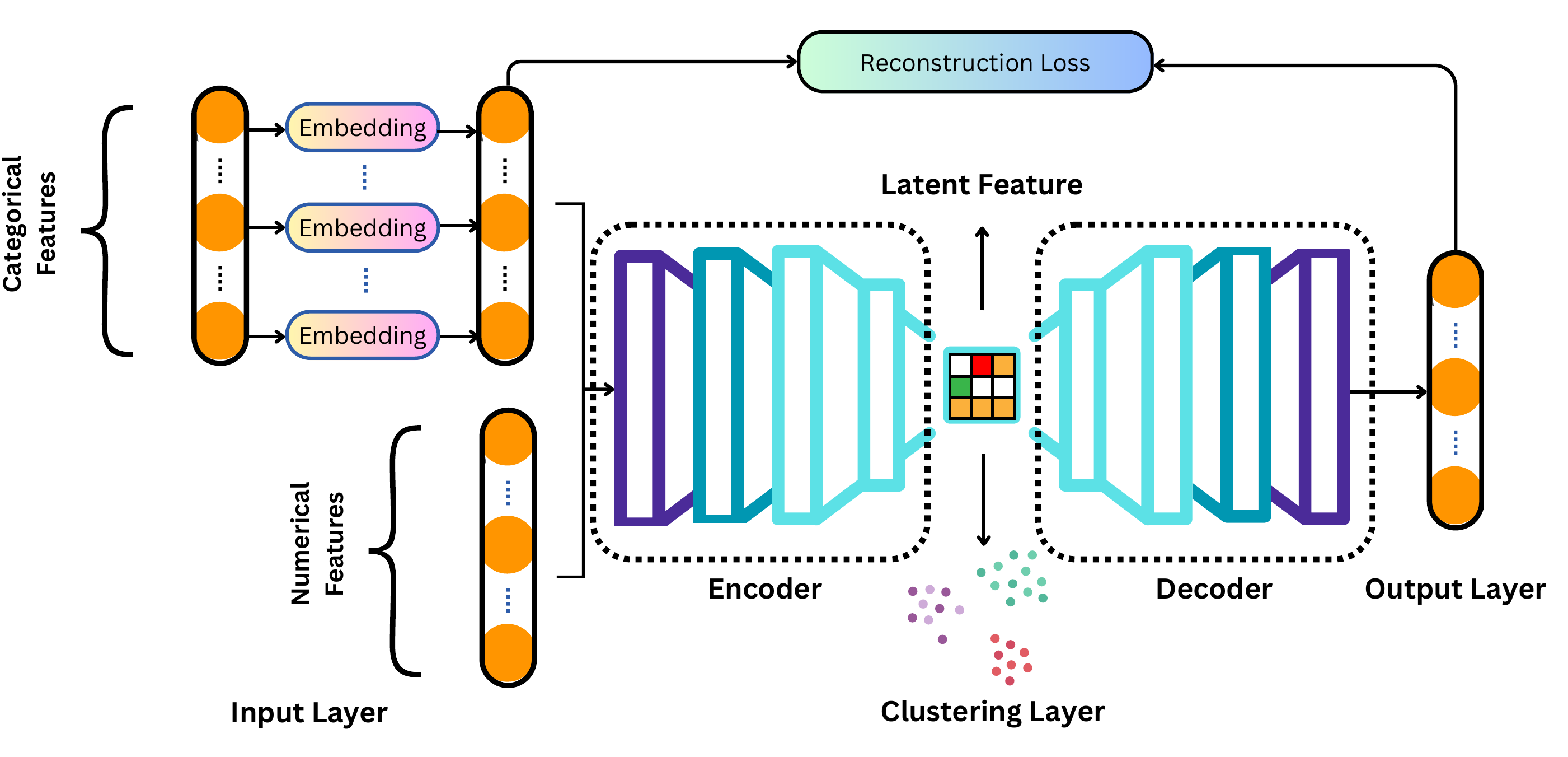} 
    \caption{DEC architecture.}
    \label{fig:DEC_Architecture} 
\end{figure}

\subsection{Deep Embedded Clustering (DEC)}
DEC is highly capable of effectively handling high-dimensional and non-linear accident data \cite{han2019short}. It utilizes deep neural networks to integrate feature learning and clustering strategies into a single framework, as illustrated in Fig. \ref{fig:DEC_Architecture}. DEC utilizes an encoder to transform input data into a lower-dimensional latent space and a clustering layer to form clusters based on the learned representations. Optimal clustering is obtained through fine-tuning the DEC hyperparameters. Careful adjustment of key parameters, such as hidden layers, latent space dimensions, and clustering coefficients, is crucial for improving the quality of clusters \cite{zhou2023identifying}. 

The DEC consists of two primary loss functions: reconstruction loss and clustering loss. The loss between input $\mathbf{x}_i$ and reconstructed output $\hat{\mathbf{x}}_i$ is minimized using reconstruction loss, which is formulated as:

\begin{equation}
\mathcal{L}_{\text{reconstruction}} = \frac{1}{N} \sum_{i=1}^{N} \| \mathbf{x}_i - \hat{\mathbf{x}}_i \|^2
\end{equation}

Clustering loss handles the divergence between estimated cluster assignment $q_{ij}$ and target distribution $p_{ij}$ through KL divergence, where N represents the number of data samples, and the number of clusters is represented by K \cite{zhou2023identifying}. 

\begin{equation}
\mathcal{L}_{\text{clustering}} = \sum_{i=1}^{N} \sum_{j=1}^{K} q_{ij} \log \frac{q_{ij}}{p_{ij}}
\end{equation}

These losses ensure that the learning is accurate by minimizing the loss during the training process and improving cluster quality.

AutoML techniques, such as Optuna, help enhance DEC performance. Optuna is a powerful tool capable of providing optimal hyperparameters by efficient searching through pruning strategies.  The primary advantage of AutoML is that it has the potential to explore a wide range of hyperparameter configurations to achieve optimal performance on the dataset. With the optimal set of hyperparameters, DEC can produce high-quality clusters, characterized by better silhouette scores, which highlights its ability to handle complex data. 

\subsection{Fundamentals of BN}

BN is a probabilistic graphical model that utilizes a directed acyclic graph (DAG), where nodes represent variables and probabilistic dependencies between nodes are represented through edges. Conditional probability tables (CPTs) are used to quantify the edges of BN \cite{kjaerulff2008bayesian}. The joint probability distribution represents conditional probability between the parent and child nodes, and its function is formulated as follows:
\begin{equation}
P(X_1, X_2, \dots, X_n) = \prod_{i=1}^n P(X_i \mid \text{Pa}(X_i))
\end{equation}

With \( X_i \) being a probabilistic variable, and \( \text{Pa}(X_i) \) representing its parent nodes \cite{kjaerulff2008bayesian}.

The BN's inference mechanism allows for the estimation of unknown variables given evidence. Bayes' theorem forms the foundation of this inference process:

\begin{equation}
P(A \mid B) = \frac{P(B \mid A) \cdot P(A)}{P(B)}
\end{equation}

The query variable and observed evidence are represented as \( A \) and \( B \) \cite{kjaerulff2008bayesian}. 

% For the case of multi-variable systems, the inference formula reads:

% \begin{equation}
% P(Q \mid E) = \frac{\sum_{H} P(Q, E, H)}{\sum_{Q, H} P(Q, E, H)}
% \end{equation}

% where \( Q \), \( E \), and \( H \) denote the query variable, observed evidence, and hidden variables, respectively.

% Computations are simplified as below by the BN (BN) to ensure conditional independence among non-connected variables and allow the latest updates with effective modeling \cite{talluri2024bayesian}.

% \begin{equation}
% P(X_i \mid \text{Pa}(X_i), \text{ND}(X_i)) = P(X_i \mid \text{Pa}(X_i))
% \end{equation}

% With \( \text{ND}(X_i) \) referring to the non-descendants of \( X_i \).

% Conditional probabilities are assigned by BN for every scenario using the advantage of obtained probabilistic relationships between the nodes. 
\begin{figure}[h]
    \centering
    \includegraphics[width=1\linewidth, height=0.23\textheight]
    {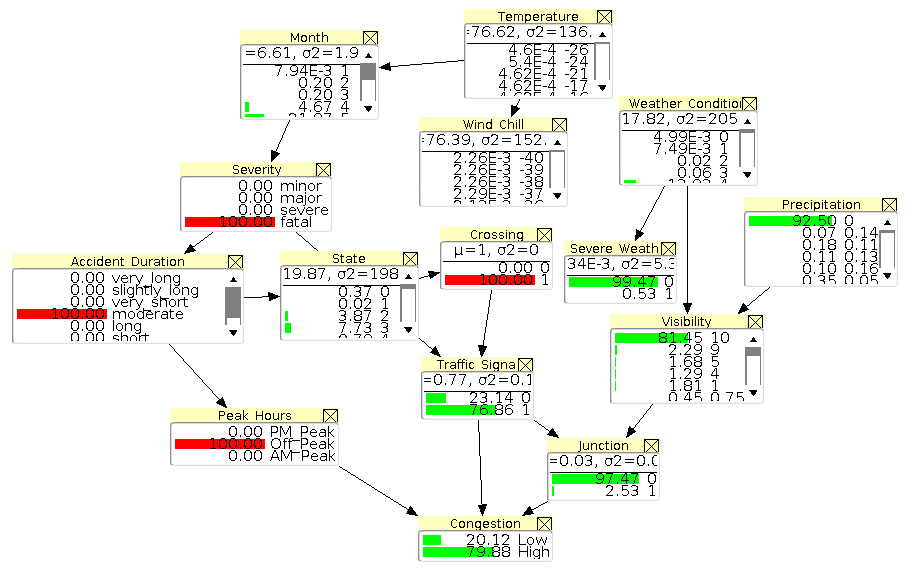} 
    \caption{Structure of proposed BN model with four evidence marked in red.}
    \label{fig:BN_with_evidence} 
\end{figure}

\section{BN Construction with SUMO Validation}

This section provides detailed information about BN modeling, followed by a description of SUMO setup and the metrics used to evaluate the BN.

\subsection{BN Construction}

The proposed BN structure consists of 14 variables, as shown in Fig. \ref{fig:BN_with_evidence}. The SHAP explainability is used to prioritize the most critical features that contribute to effective congestion prediction. For the modeling of BN, data-driven structural learning under structural causal constraints is employed. It is further enhanced by integrating the HUGIN tool and the hill-climb search algorithm to ensure a balance between model complexity and performance. This hybrid methodology effectively captures the sophisticated relationships between accident features and congestion outcomes. Furthermore, several scenarios are developed to illustrate the complex relationships between accident features and understanding the root cause.

\begin{figure}[h]
    \centering
    \includegraphics[width=0.8\linewidth]{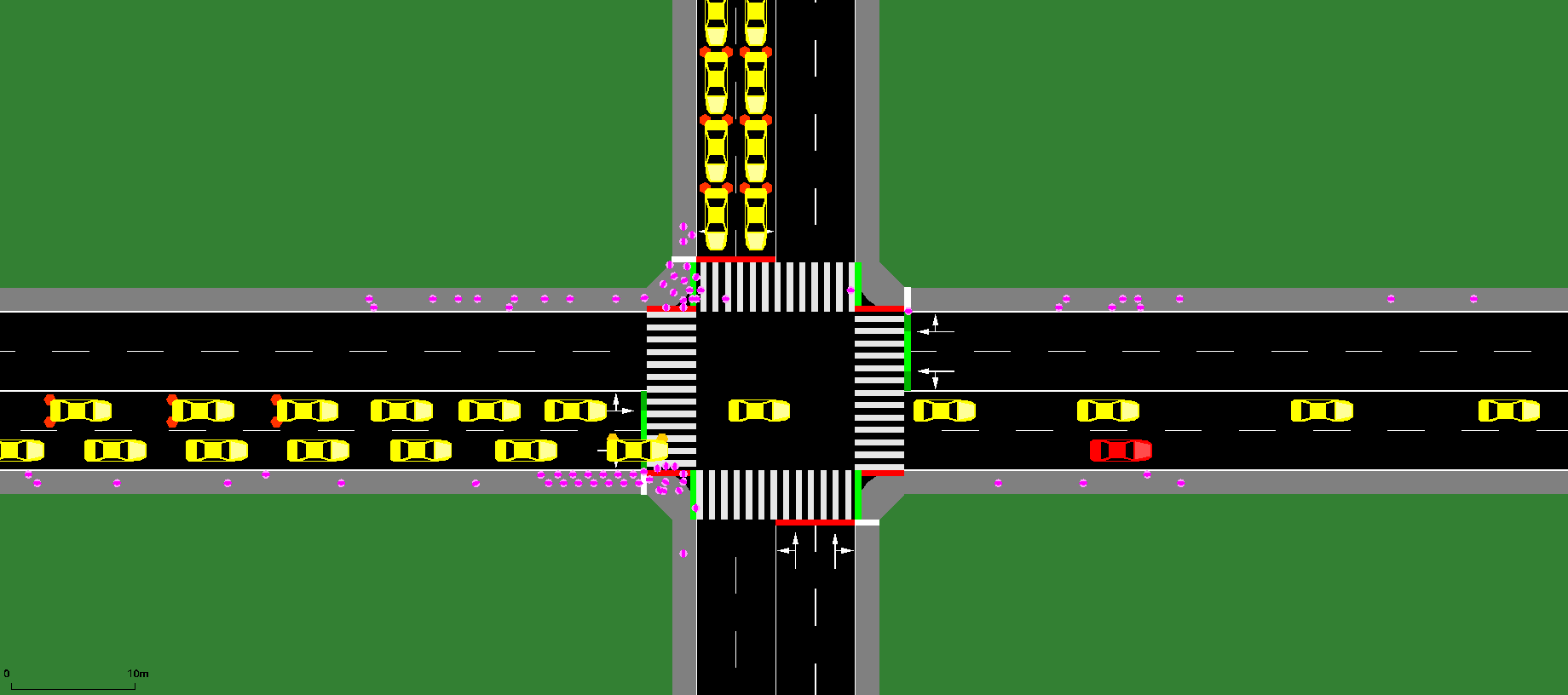} 
    \caption{Simulation of accident in SUMO. The red car is the one involved in an accident, causing traffic congestion on both arms of the depicted intersection.}
    \label{fig:SUMO_accident_sample} 
\end{figure}

\subsection{SUMO Simulation Setup}

SUMO is the most widely used robust open-source traffic simulation platform for simulating real-world traffic scenarios \cite{liu2020markov}. In this work, accident scenarios are seamlessly replicated in the SUMO environment to validate the BN’s prediction of congestion probability and analyze the impact of accidents on traffic congestion. While real-time GNSS data was not directly used, the accident dataset includes location-based features derived from reported coordinates. Crucial accident characteristics like road type, traffic signals, pedestrian crossing, accident duration, and junction information are effectively modeled in SUMO, as shown in Fig. \ref{fig:SUMO_accident_sample}. The significant elements of the SUMO network include:

\begin{itemize}
    \item Road Layout: A Road network with various intersections, crossings, and traffic signals to support a smooth flow of vehicles and pedestrians at regular intervals.
    \item Accident Scenarios: Controlled accidents are simulated by altering the speed of the vehicle drop to zero for a specific duration. Designed experiments for different accident scenarios with varying severity and junction proximity to analyze their impact on congestion characteristics.
\end{itemize}

Various traffic performance metrics, namely: Average Queue Length (AQL), Average Waiting Time (AWT), Maximum Queue Length (MQL), Average Network Speed (ANS), Queue Length (QL), Spatial Congestion Index (SCI), and Root Mean Square Error (RMSE), were recorded through SUMO simulations to quantify the accident impact and validate the correctness of BN performance. These metrics are then compared with BN congestion probability to verify the accuracy of predictions and the relevance of the framework for practical applicability in real-time.

\section{Results and Discussion}

This section discusses the significance of analysis and validations supporting the proposed framework's performance. %Initially, the temporal relationship between accidents and their impact on congestion is explored. Then, various clustering techniques are evaluated to find an optimal technique and perform effective cluster profiling. Furthermore, BN predictions are obtained for different traffic scenarios, and their correctness is validated using SUMO simulations. These structured results collectively highlight the practicality of the complete workflow in analyzing traffic congestion caused by accidents.

\subsection{Temporal Analysis of Accidents and Congestion}

\begin{figure}[h]
    \centering
    \includegraphics[width=0.8\linewidth, height=0.16\textheight]
    {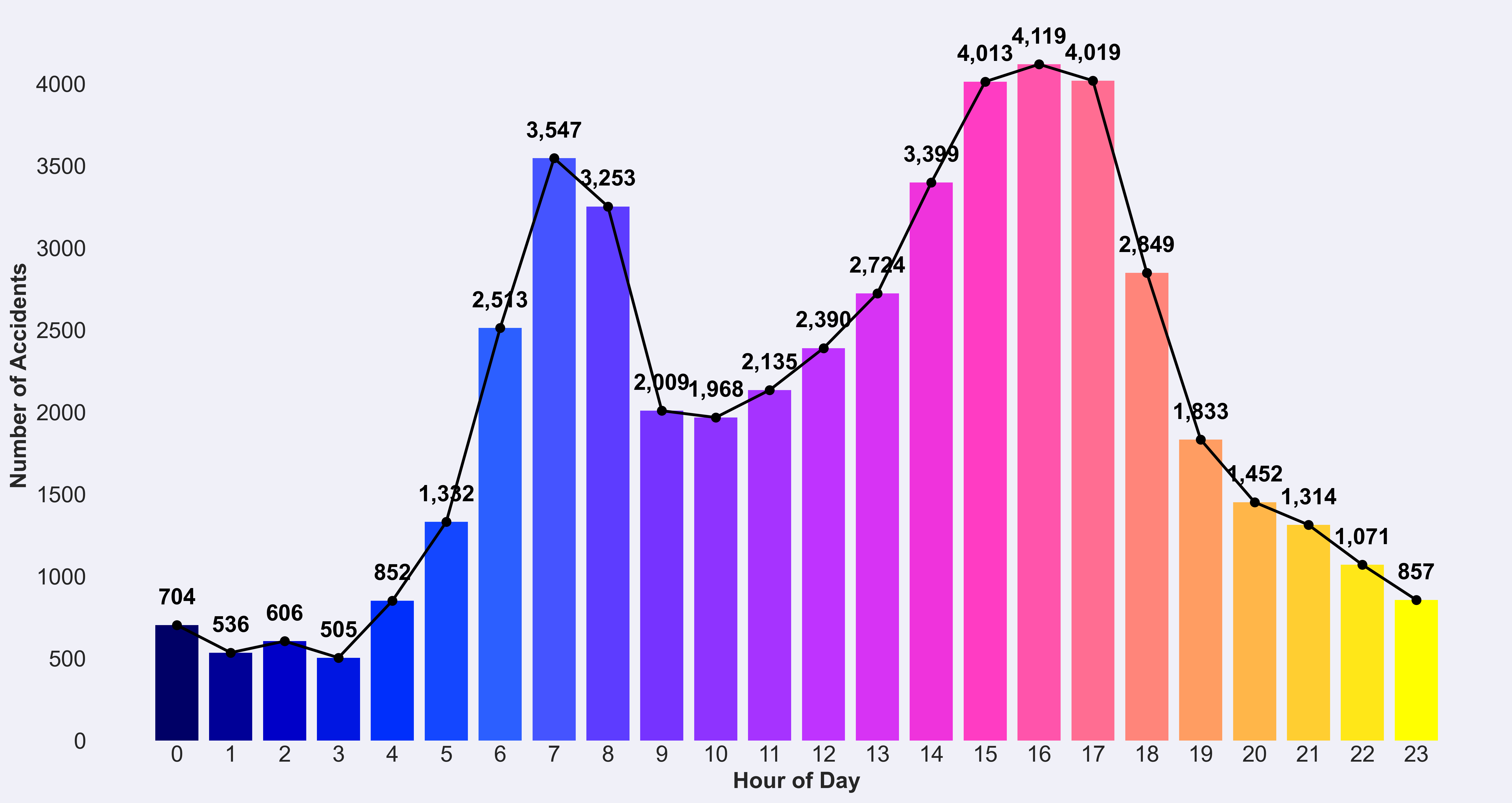} 
    \caption{Road accidents in USA on an hourly basis.}
    \label{fig:accidents_by_hour} 
\end{figure}

The likelihood of accident occurrence and its alignment with congestion peak hours is analyzed to find the patterns contributing to accidents. Fig. \ref{fig:accidents_by_hour} shows a clear pattern of lower accident counts during early morning and late night hours. Peak hours between 6-9 AM and 2-6 PM depict substantial increases in accidents with a peak of 3547 and 4119, respectively, which strongly correlates with high vehicle density during working hours. These high-risk hours underscore the need for advanced traffic mitigation strategies, which enhance road safety.

\subsection{Clustering Analysis}

DEC architecture, as shown in Fig. \ref{fig:DEC_Architecture}, consists of 11 numerical and 12 categorical features at the input layer. The encoder and decoder blocks, stacked with 190 hidden layers and feature extraction with 19 latent features, are the best set of parameters obtained from Optuna’s AutoML. The hyperparameters of DEC include a learning rate of 0.0002, a batch size of 64, and 50 epochs. Moreover, the Adam optimizer and MSE loss function are used in DEC training for optimal clustering.

\begin{table}[h]
\renewcommand{\arraystretch}{1.2} % Adjust row spacing
\centering
\caption{Silhouette Scores for Clustering Techniques Across Cluster Numbers}\label{tab:silhouette_scores}
\begin{tabular}{|l|c|c|c|c|c|}
\hline
\multirow{2}{*}{\textbf{Method}} & \multicolumn{5}{c|}{\textbf{Clusters}} \\ \cline{2-6}
                                 & \textbf{2}   & \textbf{3}   & \textbf{4}   & \textbf{5}   & \textbf{6}   \\ \hline
K-Means                          & 0.114       & 0.115       & 0.096       & 0.096       & 0.126       \\ \hline
Hierarchical                     & 0.128       & 0.137       & 0.060       & 0.062       & 0.079       \\ \hline
DBSCAN             & \multicolumn{5}{c|}{0.142 (Eps: 3.5)}              \\ \hline
DEC                         & 0.301       & 0.270       & 0.266       & 0.131       & 0.133       \\ \hline
DEC + PCA                        & 0.305       & 0.275       & 0.270       & 0.136       & 0.135       \\ \hline
DEC + SVD                        & 0.307       & 0.277       & 0.273       & 0.137       & 0.153       \\ \hline
DEC + AutoML                     & \textbf{0.490} & 0.484    & 0.469       & 0.300       & 0.194       \\ \hline
\end{tabular}
\end{table}

The silhouette score across various cluster configurations explains the effectiveness of different clustering techniques. Table \ref{tab:silhouette_scores} shows that traditional methods like k-means, DBSCAN, and hierarchical techniques fail to accurately capture the complex relationships of accident features and exhibit a very low score. K-means achieved a maximum value of 0.1255 with 6 clusters, and hierarchical clustering showed a peak of 0.1372 during 3 clusters. On the contrary, DBSCAN yielded a maximum silhouette score of 0.1416, which corresponds to the optimal configuration of Epsilon, set to 3.5. The advanced clustering method, DEC, outperformed traditional techniques with a better score. Baseline DEC gave a maximum result of 0.3007 at 2 clusters. The performance of baseline DEC can be seen to improve further when PCA and SVD dimensionality reduction techniques are used. AutoML optimization is incorporated into DEC to achieve the best score of 0.4902 with 2 clusters (cluster 0 and cluster 1). AutoML enhances DEC's capability to identify meaningful patterns in the dataset by recommending the optimal set of DEC parameters based on the input data. The analysis highlights the importance of using advanced clustering techniques to automate and extract meaningful insights from the intricate dataset.

\subsection{Cluster Profiling and Feature Explainability Using SHAP Analysis}

\begin{figure}[h]
    \centering
    \includegraphics[width=1\linewidth, height=0.3\textheight]{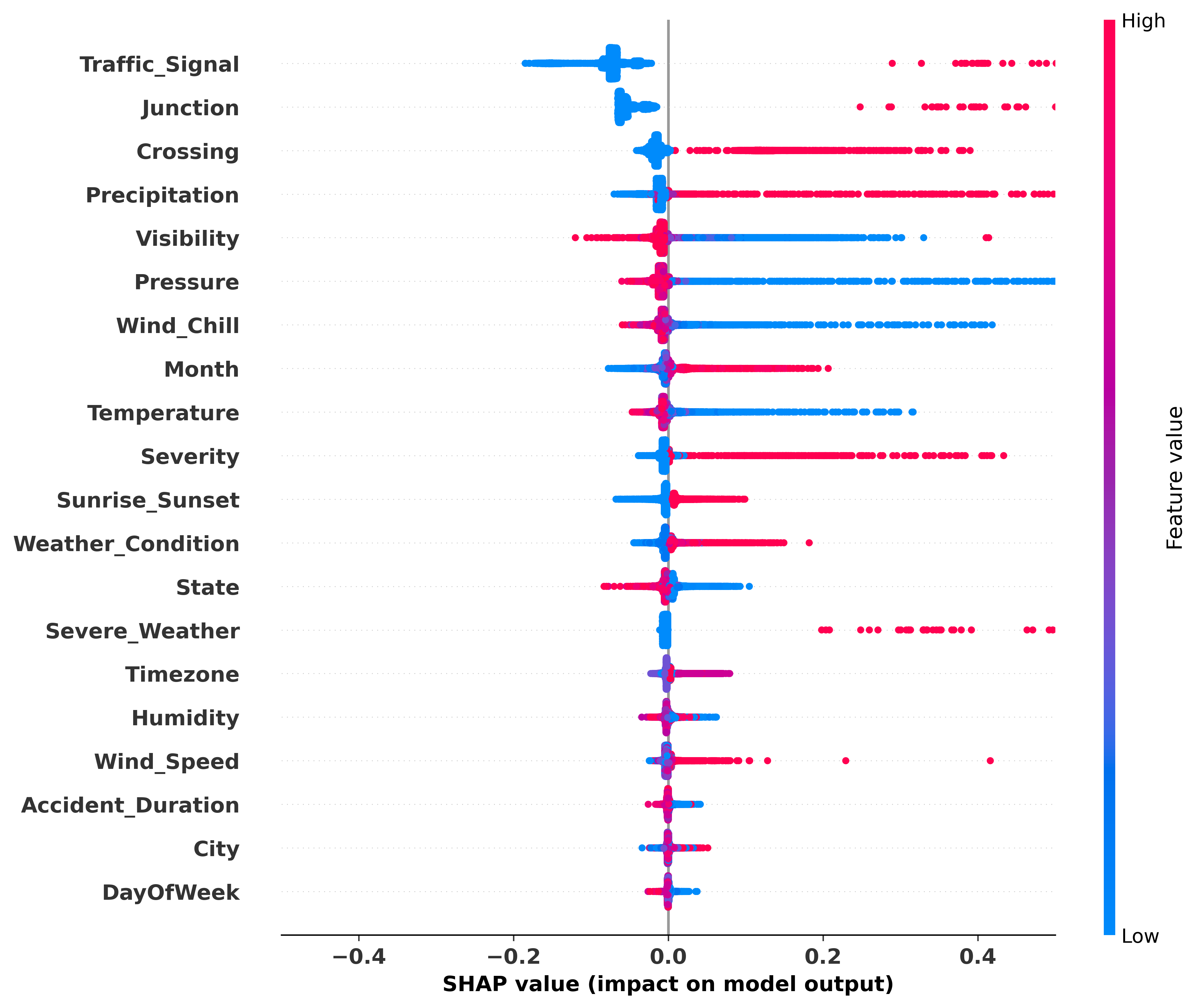} 
    \caption{SHAP values impact on model output for cluster 0.}
    \label{fig:figure_shapely} 
\end{figure}

Advanced and explainable techniques like SHAP perform detailed cluster profiling to analyze and interpret the features with the highest contribution and aids in assigning congestion labels (Low or High) to the 2 clusters obtained through the DEC + AutoML method. Interpreting deep latent features posed a challenge due to their abstract nature. SHAP interpretability trade-offs are addressed by linking cluster assignments back to original input features to retain interpretability.

The Fig. \ref{fig:figure_shapely} shows the SHAP plot with all the key features like traffic signals, junctions, and crossings based on their contribution to cluster 0. The feature values are shown in Fig. \ref{fig:figure_shapely}, with blue points indicating low values and red points indicating high feature values. The SHAP value on the horizontal axis shows the impact of each feature on the model output. For instance, traffic signals exhibit negative SHAP values, as they negatively impact congestion for cluster 0; however, features such as precipitation, crossing, and severity have a positive impact on congestion for cluster 0.	 

% \begin{figure}[h]
%     \centering
%     \includegraphics[width=0.7\linewidth, height=0.18\textheight]
%     {figures/cluster_radar_c.png} 
%     \caption{Illustrating the cluster profiling using radar plot.}
%     \label{fig:cluster_radar} 
% \end{figure}

% Top contributing features from SHAP analysis are used for cluster profiling with the help of a radar plot, as shown in Figure \ref{fig:cluster_radar}. The profiling process uses the top five features, namely traffic signal, junction, crossing, precipitation, and visibility, to understand the hidden patterns and relationships differentiating the clusters. Furthermore, radar plots using these five variables deepen feature understanding, which helps in effective cluster profiling for assigning meaningful labels to the identified clusters, such as “Low Congestion” or “High Congestion”. 

Cluster 0 with high positive SHAP values, as seen in Fig. \ref{fig:figure_shapely}, of features like traffic signal, junction, crossing, precipitation, severity, and severe weather are labeled as High Congestion as they cause significant traffic delays and bottlenecks. On the contrary, the Low Congestion label is provided to cluster 1, which has a high negative impact on crucial features, as the traffic scenarios have smoother traffic flow. The SHAP-based profiling ensures that assigned labels accurately represent traffic congestion states, thereby making them closer to real-world traffic conditions. SHAP values can help traffic management identify key accident factors influencing congestion, enabling targeted interventions, such as adjusting signal timings or deploying resources, based on dominant features like severity or proximity to junctions.

\begin{table}[h!]
\centering
\renewcommand{\arraystretch}{0.8}
\caption{Bayesian Network Model Evaluation Results}
\begin{tabular}{lcc}
\toprule
\multicolumn{3}{l}{\textbf{Overall Metrics}} \\
\midrule
\textbf{Metric}        & \textbf{Value}      \\
\midrule
Overall Accuracy       & 0.9564             \\
Sensitivity (True Positive Rate) & 0.9902  \\
Specificity (True Negative Rate) & 0.8297  \\
\midrule
\multicolumn{3}{l}{\textbf{Class-wise Metrics}} \\
\midrule
\textbf{Class}         & \textbf{Metric}    & \textbf{Value}      \\
\midrule
\multirow{3}{*}{Low Congestion} & Precision    & 0.9578             \\
                              & Recall       & 0.8297             \\
                              & F1 Score     & 0.8892             \\
\midrule
\multirow{3}{*}{High Congestion} & Precision    & 0.9561             \\
                              & Recall       & 0.9902             \\
                              & F1 Score     & 0.9729             \\

\bottomrule
\end{tabular}

\label{tab:bn_evaluation}
\end{table}

\begin{table}[!h]
\centering
\renewcommand{\arraystretch}{1.3}
\caption{Demonstrating the impact of variables on congestion using different scenarios}\label{tab:scenarios_tab_new}
\begin{tabular}{|c|l|l|}
\hline
\textbf{Scenarios} & \textbf{Variable (state)} & \textbf{Congestion} \\ \hline

1 & \begin{tabular}[c]{@{}l@{}}Severity (\textbf{Minor}), Crossing (Yes), \\ Peak\_Hours (OFF Peak), \\ Accident\_Duration (moderate)\end{tabular} & 
\begin{tabular}[c]{@{}l@{}}Low (51.92\%), \\ High (48.08\%)\end{tabular} \\ \hline

2 & \begin{tabular}[c]{@{}l@{}}Severity (\textbf{Fatal}), Crossing (Yes), \\ Peak\_Hours (OFF Peak), \\ Accident\_Duration (moderate)\end{tabular} & 
\begin{tabular}[c]{@{}l@{}}Low (20.12\%), \\ High (79.88\%)\end{tabular} \\ \hline

3 & \begin{tabular}[c]{@{}l@{}}Junction (\textbf{No}), Crossing (Yes), \\ Peak\_Hours (AM Peak), \\ Accident\_Duration (very short)\end{tabular} & 
\begin{tabular}[c]{@{}l@{}}Low (51.74\%), \\ High (48.26\%)\end{tabular} \\ \hline

4 & \begin{tabular}[c]{@{}l@{}}Junction (\textbf{Yes}), Crossing (Yes), \\ Peak\_Hours (AM Peak), \\ Accident\_Duration (very short)\end{tabular} & 
\begin{tabular}[c]{@{}l@{}}Low (1.88\%), \\ High (98.12\%)\end{tabular} \\ \hline

\end{tabular}
\end{table}

% \begin{table*}[h]
% \centering
% \renewcommand{\arraystretch}{1.2} % Adjust row spacing
% \caption{Validation of Bayesian Network Predictions with SUMO Metrics}
% \begin{tabular}{|c|c|c|c|c|}
% \hline
% \textbf{Metric} & \textbf{Scenario 1} & \textbf{Scenario 2} & \textbf{Scenario 3} & \textbf{Scenario 4} \\ \hline

% \multicolumn{5}{|c|}{\textbf{SUMO Simulation}} \\ \hline

% \textbf{Avg} & 
% 4.6 & 8.2 & 4.6 & 4.3 \\ \hline

% \textbf{Avg.} & 
% 42.8 & 64.6 & 42.6 & 89.3 \\ \hline

% \textbf{Max. } & 
% 6 & 16 & 6 & 14 \\ \hline

% \textbf{Avg. } & 
% 3.22 & 2.57 & 3.32 & 1.89 \\ \hline

% \textbf{Queue } & 
% 45 & 75 & 40 & 120 \\ \hline

% \textbf{Spa} & 
% 0.35 & 0.65 & 0.30 & 0.85 \\ \hline

% \textbf{RMSE} & 
% 0.047 & 0.052 & 0.038 & 0.029 \\ \hline

% % \multirow{2}{*}{\textbf{BN Prediction}} & 
% % Low: 51.96\% & Low: 20.12\% & Low: 42.99\% & Low: 1.88\% \\ 
% %  & High: 48.04\% & High: 79.88\% & High: 57.01\% & High: 98.12\% \\ \hline

% \multicolumn{5}{|c|}{\textbf{BN Model}} \\ \hline

% \textbf{BN } & 
% (51.96\%)  &  (79.88\%) &  (57.01\%) &  (98.12\%) \\ \hline

% \end{tabular}
% \label{tab:bn_validation}
% \end{table*}
\begin{table*}[h]
\centering
\renewcommand{\arraystretch}{1.2} % Adjust row spacing
\caption{Validation of Bayesian Network Predictions with SUMO Metrics}
\begin{tabular}{|c|c|c|c|c|}
\hline
\textbf{Metric} & \textbf{Scenario 1} & \textbf{Scenario 2} & \textbf{Scenario 3} & \textbf{Scenario 4} \\ \hline

\multicolumn{5}{|c|}{\textbf{SUMO Simulation}} \\ \hline

\textbf{Average Queue Length} & 
4.6 & 8.2 & 4.6 & 4.3 \\ \hline

\textbf{Average Waiting Time (s)} & 
42.8 & 64.6 & 42.6 & 89.3 \\ \hline

\textbf{Maximum Queue Length} & 
6 & 16 & 6 & 14 \\ \hline

\textbf{Average Network Speed (m/s)} & 
3.22 & 2.57 & 3.32 & 1.89 \\ \hline

\textbf{Queue Length (m)} & 
45 & 75 & 40 & 120 \\ \hline

\textbf{Spatial Congestion Index} & 
0.35 & 0.65 & 0.30 & 0.85 \\ \hline

\textbf{Root Mean Square Error} & 
0.047 & 0.052 & 0.038 & 0.029 \\ \hline

% \multirow{2}{*}{\textbf{BN Prediction}} & 
% Low: 51.96\% & Low: 20.12\% & Low: 42.99\% & Low: 1.88\% \\ 
%  & High: 48.04\% & High: 79.88\% & High: 57.01\% & High: 98.12\% \\ \hline

\multicolumn{5}{|c|}{\textbf{BN Model}} \\ \hline

\textbf{BN Congestion Prediction} & 
Low (51.96\%)  & High (79.88\%) & High (57.01\%) & High (98.12\%) \\ \hline

\end{tabular}
\label{tab:bn_validation}
\end{table*}

The practical assignment of congestion labels enhances the interpretability of clustering. This step is crucial before BN prediction, as congestion serves as a target label. These congestion labels form probabilistic relationships between accident features and congestion levels, promoting proactive insights and building an efficient BN model.

\subsection{BN Predictions and Scenarios Evaluations}

Table \ref{tab:bn_evaluation} shows the BN model's reliability and accuracy in predicting congestion probability. The proposed BN model achieved a remarkable accuracy of 95.64\% and a true positive rate of 99.02\% because it perfectly predicts high congestion instances. 82.97\% of the true negative rate shows that low congestion scenarios are identified precisely to reduce false positives. BN models' precession, recall, and F1 scores are featured through class-wise evaluation. The case of low congestion achieved an F1 score of 88.92\%, while high congestion showed a 97.29\% F1 score, depicting its definite estimation capability of serious high-congestion events.

As shown in Table \ref{tab:scenarios_tab_new}, four different accident scenarios are created to determine the BN model's predictive potential. The primary purpose of these scenarios is to illustrate the impact of accident features on congestion states. For instance, in Scenario 1, four evidence of accidents are given; if the accident occurred at a crossing during off-peak hours and the severity of the accident is minor where the accident duration is moderate, then the probability of this accident causing low congestion is 51.92\%. Similarly, scenario 2 is created using the same evidence as scenario 1, but with the assumption that the severity of the accident is fatal. Then, there is a high likelihood of 79.88\%, causing high congestion. Based on scenarios 1 and 2, it is clear that the severity of the accident has a significant impact on congestion. Similarly, scenarios 3 and 4 show that accidents occurring near a junction has a substantial effect, with a 98.12\% chance of causing high congestion.

These results help researchers verify the effectiveness of the BN framework and plan advanced traffic management strategies more effectively. Proactive decision-making through precise quantification of congestion probabilities helps mitigate congestion caused by uncertainties, such as accidents. The scenario-specific estimates underscore the framework's practical significance and enhance its reliability for use in real-world transportation networks, particularly when validated using traffic simulators such as SUMO.

\subsection{SUMO Validation}

SUMO simulation is used to replicate the accident that occurred, as illustrated in Table \ref{tab:scenarios_tab_new}, and to observe the level of congestion caused. This process will ensure the accuracy and validity of proposed BN models in predicting congestion based on information about the accident. Moreover, additional evidence about the accidents could further strengthen the BN model's prediction by increasing the likelihood of the congestion state. For this purpose, SUMO network parameters are carefully configured to reflect similar evidence as stated in Table \ref{tab:scenarios_tab_new} and simulated 4 accidents accordingly.

For instance, in scenario 4, with a significant accident occurring near the intersection, the probability of congestion is very high, with 98.12\% due to extreme traffic conditions. This scenario was simulated by setting the accident to occur at 600 seconds during peak hours, ensuring a high congestion impact. The vehicle's length after the accident is set to 80 meters, and the speed is reduced to 0 m/s to obtain a large-scale blocking and standstill situation, respectively. Additional delays are introduced by adding pedestrian movements across the lanes. With these parameter settings, the results, as shown in Table \ref{tab:bn_validation}, indicate that the highest weighting time of 89.3 seconds, the queue length of 120 meters, and the queue distribution of 13 to 14 vehicles on the east were observed. A detailed analysis of all the metrics obtained from the SUMO simulation showed a close match between the resulting congestion and BN’s congestion prediction.  

Further examination of Table \ref{tab:bn_validation} shows that in scenario 3, the probability of high congestion is 57.01\%.  SUMO simulation shows a lower waiting time of 42.6 seconds and a minimal queue length of 40 meters, reflecting the moderate impact of accidents on traffic conditions. On the contrary, scenario 1 shows a low probability of congestion occurrence with a smaller queue length of 5 cars and a waiting time of 42.8 seconds. This comparative analysis demonstrates that BN’s predictions strongly correlate with simulated traffic behaviors, verifying its performance and interpretability under various accident scenarios. The reliability of the framework improved due to the use of this evidence-based method. Severe and non-severe congestion scenarios were effectively evaluated using the SUMO environment to analyze and build modern and efficient traffic management strategies.

\begin{figure}[h]
    \centering
    \includegraphics[width=0.9\linewidth]{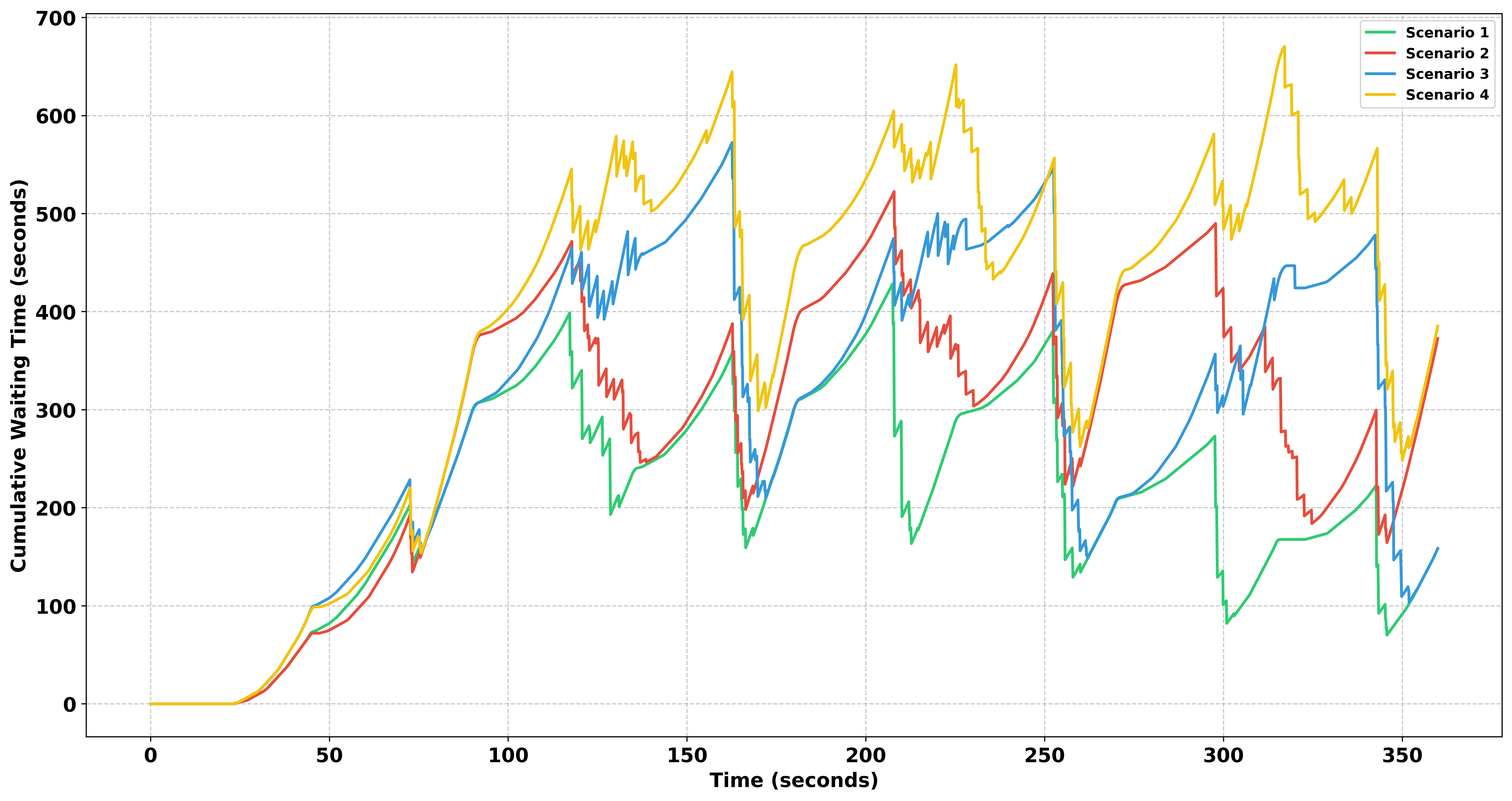} 
    \caption{Waiting time for different scenarios obtained from SUMO.}
    \label{fig:waiting_time} 
\end{figure}

Furthermore, the waiting time plot is shown in Fig. \ref{fig:waiting_time} to demonstrate the SUMO results visually. This plot presents a comprehensive analysis of traffic congestion across all four scenarios, as simulated, in terms of cumulative waiting times for each scenario. Scenario 4, which faces challenging conditions and the highest possibility of congestion during an accident, exhibits a steep increase in cumulative waiting time. Scenario 1 has the lowest waiting time due to minimum congestion circumstances. Intermediate trends are observed in scenarios 2 and 3 due to moderate congestion probability. This visualization enhances the reliability of BN’s performance by clearly differentiating between low and high congestion scenarios in terms of accident characteristics.

\section{Conclusion}

This paper proposes a robust framework for assigning the congestion labels from the accident data using an advanced clustering technique, namely Deep embedding clustering, whose performance is enhanced by an AutoML optimizer. Furthermore, the BN model is developed using the SHAP-based feature extraction. Results demonstrate that DEC outperforms traditional clustering techniques, achieving the highest silhouette score of 0.49. The performance of the BN model for congestion prediction achieved 95.6\% accuracy, highlighting its ability to understand the relationship between accident characteristics and congestion outcomes effectively. Moreover, the proposed BN model's interpretability and explainability are illustrated through scenario evaluation. Finally, SUMO simulation is used to create realistic accident scenarios based on evidence from the BN model, thereby replicating real-world traffic environments and evaluating the correctness of predictions obtained from the BN model. The evaluation shows that SUMO metrics and BN prediction are closely aligned. 

The limitations of this study are regional bias due to the U.S.-based accident data and simplified assumptions and static layouts in SUMO simulations. We used the silhouette score due to its effectiveness in measuring both cohesion and separation without requiring labeled data; however, we agree that additional metrics, such as Davies-Bouldin or Calinski-Harabasz, can enrich future evaluations. The current scope of this work is to validate the predictive capabilities of the proposed BN model. This framework lays a foundation for using explainable BN for effective traffic management. In the future, we plan to explore the integration of real-time sensor data to apply adaptive traffic management strategies, such as enhancing the reinforcement learning agent to optimize traffic light control, and evaluating the model’s transferability to diverse urban contexts for improving public safety, emergency vehicle routing, and urban mobility.

\bibliographystyle{IEEEtran}
\bibliography{references}

\begin{thebibliography}{10}
\providecommand{\url}[1]{#1}
\csname url@samestyle\endcsname
\providecommand{\newblock}{\relax}
\providecommand{\bibinfo}[2]{#2}
\providecommand{\BIBentrySTDinterwordspacing}{\spaceskip=0pt\relax}
\providecommand{\BIBentryALTinterwordstretchfactor}{4}
\providecommand{\BIBentryALTinterwordspacing}{\spaceskip=\fontdimen2\font plus
\BIBentryALTinterwordstretchfactor\fontdimen3\font minus \fontdimen4\font\relax}
\providecommand{\BIBforeignlanguage}[2]{{%
\expandafter\ifx\csname l@#1\endcsname\relax
\typeout{** WARNING: IEEEtran.bst: No hyphenation pattern has been}%
\typeout{** loaded for the language `#1'. Using the pattern for}%
\typeout{** the default language instead.}%
\else
\language=\csname l@#1\endcsname
\fi
#2}}
\providecommand{\BIBdecl}{\relax}
\BIBdecl

\bibitem{santos2021machine}
D.~Santos, J.~Saias, P.~Quaresma, and V.~B. Nogueira, ``Machine learning approaches to traffic accident analysis and hotspot prediction,'' \emph{Computers}, vol.~10, no.~12, p. 157, 2021.

\bibitem{afrin2021probabilistic}
T.~Afrin and N.~Yodo, ``A probabilistic estimation of traffic congestion using bayesian network,'' \emph{Measurement}, vol. 174, p. 109051, 2021.

\bibitem{luan2022traffic}
S.~Luan, R.~Ke, Z.~Huang, and X.~Ma, ``Traffic congestion propagation inference using dynamic bayesian graph convolution network,'' \emph{Transportation research part C: emerging technologies}, vol. 135, p. 103526, 2022.

\bibitem{talluri2024bayesian}
K.~K. Talluri and G.~Weidl, ``Bayesian network for analysis and prediction of traffic congestion using the accident data.'' in \emph{VEHITS}, 2024, pp. 19--30.

\bibitem{mudgil2022intensity}
P.~Mudgil and I.~Joshi, ``Intensity of traffic due to road accidents in us: A predictive model,'' in \emph{International Conference on Innovative Computing and Communications: Proceedings of ICICC 2021, Volume 3}.\hskip 1em plus 0.5em minus 0.4em\relax Springer, 2022, pp. 43--50.

\bibitem{ManzoorRFCNN2021}
M.~Manzoor, M.~Umer, S.~Sadiq, A.~Ishaq, S.~Ullah, H.~A. Madni, and C.~Bisogni, ``Rfcnn: Traffic accident severity prediction based on decision level fusion of machine and deep learning model,'' \emph{IEEE Access}, vol.~9, pp. 128\,359--128\,371, 2021.

\bibitem{islam2021clustering}
M.~R. Islam, I.~J. Jenny, M.~Nayon, M.~R. Islam, M.~Amiruzzaman, and M.~Abdullah-Al-Wadud, ``Clustering algorithms to analyze the road traffic crashes,'' in \emph{2021 International Conference on Science \& Contemporary Technologies (ICSCT)}.\hskip 1em plus 0.5em minus 0.4em\relax IEEE, 2021, pp. 1--6.

\bibitem{zhou2023identifying}
R.~Zhou, H.~Huang, J.~Lee, X.~Huang, J.~Chen, and H.~Zhou, ``Identifying typical pre-crash scenarios based on in-depth crash data with deep embedded clustering for autonomous vehicle safety testing,'' \emph{Accident Analysis \& Prevention}, vol. 191, p. 107218, 2023.

\bibitem{desai2023optimal}
D.~D. Desai, J.~Dey, S.~K. Satapathy, S.~Mishra, S.~N. Mohanty, P.~Mishra, and S.~K. Panda, ``Optimal ambulance positioning for road accidents with deep embedded clustering,'' \emph{IEEE Access}, vol.~11, pp. 59\,917--59\,934, 2023.

\bibitem{zheng2020determinants}
Z.~Zheng, Z.~Wang, L.~Zhu, and H.~Jiang, ``Determinants of the congestion caused by a traffic accident in urban road networks,'' \emph{Accident Analysis \& Prevention}, vol. 136, p. 105327, 2020.

\bibitem{zheng2024exploring}
Z.~Zheng, Z.~Wang, S.~Liu, and W.~Ma, ``Exploring the spatial effects on the level of congestion caused by traffic accidents in urban road networks: A case study of beijing,'' \emph{Travel behaviour and society}, vol.~35, p. 100728, 2024.

\bibitem{talluri2024impact}
K.~K. Talluri and G.~Weidl, ``Impact of accidents on traffic congestions: A bayesian network approach using real city data,'' in \emph{Emerging Cutting-Edge Applied Research and Development in Intelligent Traffic and Transportation Systems}.\hskip 1em plus 0.5em minus 0.4em\relax IOS Press, 2024, pp. 64--78.

\bibitem{kamh2024exploring}
H.~Kamh, S.~H. Alyami, A.~Khattak, M.~Alyami, and H.~Almujibah, ``Exploring road traffic accidents hotspots using clustering algorithms and gis-based spatial analysis,'' \emph{IEEE Access}, 2024.

\bibitem{han2019short}
L.~Han, K.~Zheng, L.~Zhao, X.~Wang, and X.~Shen, ``Short-term traffic prediction based on deepcluster in large-scale road networks,'' \emph{IEEE Transactions on Vehicular Technology}, vol.~68, no.~12, pp. 12\,301--12\,313, 2019.

\bibitem{bazzi2025upper}
A.~Bazzi, R.~Bomfin, M.~Mezzavilla, S.~Rangan, T.~Rappaport, and M.~Chafii, ``Upper mid-band spectrum for 6g: Vision, opportunity and challenges,'' \emph{arXiv preprint arXiv:2502.17914}, 2025.

\bibitem{moosavi2019accident}
S.~Moosavi, M.~H. Samavatian, S.~Parthasarathy, R.~Teodorescu, and R.~Ramnath, ``Accident risk prediction based on heterogeneous sparse data: New dataset and insights,'' in \emph{Proceedings of the 27th ACM SIGSPATIAL international conference on advances in geographic information systems}, 2019, pp. 33--42.

\bibitem{kjaerulff2008bayesian}
U.~B. Kjaerulff and A.~L. Madsen, ``Bayesian networks and influence diagrams,'' \emph{Springer Science+ Business Media}, vol. 200, p. 114, 2008.

\bibitem{liu2020markov}
S.~Liu, S.~Lin, Y.~Wang, B.~De~Schutter, and W.~Lam, ``A markov traffic model for signalized traffic networks based on bayesian estimation,'' \emph{IFAC-PapersOnLine}, vol.~53, no.~2, pp. 15\,029--15\,034, 2020.

\end{thebibliography}

\end{document}